\documentclass[journal]{IEEEtai}

\usepackage[colorlinks,urlcolor=blue,linkcolor=blue,citecolor=blue]{hyperref}

\usepackage{color,array}

\usepackage{graphicx}
\usepackage{makecell}
\usepackage{xcolor}
\usepackage{orcidlink}
\usepackage{multirow}
\usepackage{url}

\begin{document}

\title{Neurosymbolic Reinforcement Learning and Planning: A Survey}

\author{Kamal Acharya\orcidlink{0000-0002-9712-0265}, \IEEEmembership{Graduate Student Member, IEEE}, Waleed Raza\orcidlink{0000-0002-4142-3195} ,\IEEEmembership{Graduate Student Member, IEEE}, Carlos Dourado\orcidlink{0000-0002-5137-1018}, \IEEEmembership{Member, IEEE}, Alvaro Velasquez\orcidlink{https://orcid.org}, \IEEEmembership{Member, IEEE} and Houbing Herbert Song\orcidlink{0000-0003-2631-9223}, \IEEEmembership{Fellow, IEEE}
\thanks{Manuscript received March 10, 2023. This work was supported in part by the U.S. National Science Foundation under Grant No. 2309760 and Grant No. 2317117.}
\thanks{K. Acharya and H. Song are with the Department of Information Systems, University of Maryland, Baltimore County, Baltimore, MD 21250 USA (e-mail: kamala2@umbc.edu; songh@umbc.edu).}
\thanks{W. Raza is with the Department of Electrical Engineering and Computer Science, Embry-Riddle Aeronautical University, Daytona Beach, FL 32114 USA (e-mail: razaw@my.erau.edu).}
\thanks{C. Dourado is with the Department of Electrical Engineering and Computer Science, Embry-Riddle Aeronautical University, Daytona Beach, FL 32114 USA (e-mail: douradoc@erau.edu).}
\thanks{A. Velasquez is with the Department of Computer Science, University of Colorado, Boulder, CO 80309 USA (e-mail: alvaro.velasquez@colorado.edu).}}

\markboth{IEEE Transactions on Artificial Intelligence, Vol. 00, No. 0, Month 2020}
{Acharya \MakeLowercase{\textit{et al.}}: Neurosymbolic Reinforcement Learning and Planning: A Survey}

\maketitle

\begin{abstract}
The area of Neurosymbolic Artificial Intelligence (Neurosymbolic AI) is rapidly developing and has become a popular research topic, encompassing sub-fields such as Neurosymbolic Deep Learning (Neurosymbolic DL) and Neurosymbolic Reinforcement Learning (Neurosymbolic RL). Compared to traditional learning methods, Neurosymbolic AI offers significant advantages by simplifying complexity and providing transparency and explainability. Reinforcement Learning(RL), a long-standing Artificial Intelligence(AI) concept that mimics human behavior using rewards and punishment, is a fundamental component of Neurosymbolic RL, a recent integration of the two fields that has yielded promising results. The aim of this paper is to contribute to the emerging field of Neurosymbolic RL by conducting a literature survey. Our evaluation focuses on the three components that constitute Neurosymbolic RL: neural, symbolic, and RL. We categorize works based on the role played by the neural and symbolic parts in RL, into three taxonomies:Learning for Reasoning, Reasoning for Learning and Learning-Reasoning. These categories are further divided into sub-categories based on their applications. Furthermore, we analyze the RL components of each research work, including the state space, action space, policy module, and RL algorithm. Additionally, we identify research opportunities and challenges in various applications within this dynamic field.
\end{abstract}
\begin{IEEEImpStatement}
Neurosymbolic RL has captured the interest of both the academic and industrial communities, as researchers strive to develop a reliable and robust model capable of achieving practical performance. Despite this, there is a lack of a comprehensive documented survey that delves into and scrutinizes the field of Neurosymbolic RL as a whole. While several survey papers devoted to Neurosymbolic AI and many more concerning RL are available, there has been no noteworthy contribution that surveys the intersection of these areas. As a result, the purpose of this article is to bridge this gap by presenting a broad range of relevant papers that have been published, with a focus on the three main elements of Neurosymbolic RL: neural, symbolic, and RL. The article conducts an analysis, identifies potential research opportunities, along with the challenges.
\end{IEEEImpStatement}


\section{Introduction}

\IEEEPARstart{N}{eurosymbolic} Artificial Intelligence (Neurosymbolic AI), a budding field of Artificial Intelligence(AI), has garnered significant attention in recent times as it combines both neural and symbolic traditions to enhance the performance of neural network models. In this context, the term "neural" pertains to neural network primarily, while "symbolic" refers to the use of various mathematical logic and algorithms for symbolic manipulation. Reinforcement Learning (RL), another emerging area of machine learning, revolves around agents operating in various environments to maximize their rewards. It dates back to the early days of cybernetics and has gained rapid interest in the machine learning and AI communities over the last five to ten years. RL involves programming agents by reward and punishment without specifying how to accomplish the task, and it encompasses statistics, psychology, neuroscience, and computer science. However, there are significant computational challenges to overcome\cite{kaelbling1996reinforcement}. Deep Reinforcement Learning (DRL), which replaces tabular methods of estimating state values with function approximation, has eliminated the need to store all state value pairs in the table, enabling the agent to generalize the value of states that it has never encountered before. DRL has been utilized in programs that have defeated the best human players in game of Go\cite{silver2016mastering}. Additionally, an AI agent named AlphaStar\cite{liu2021introduction} beat the world's best StarCraft II player.

\begin{figure}
\centerline{\includegraphics[width=18.5pc]{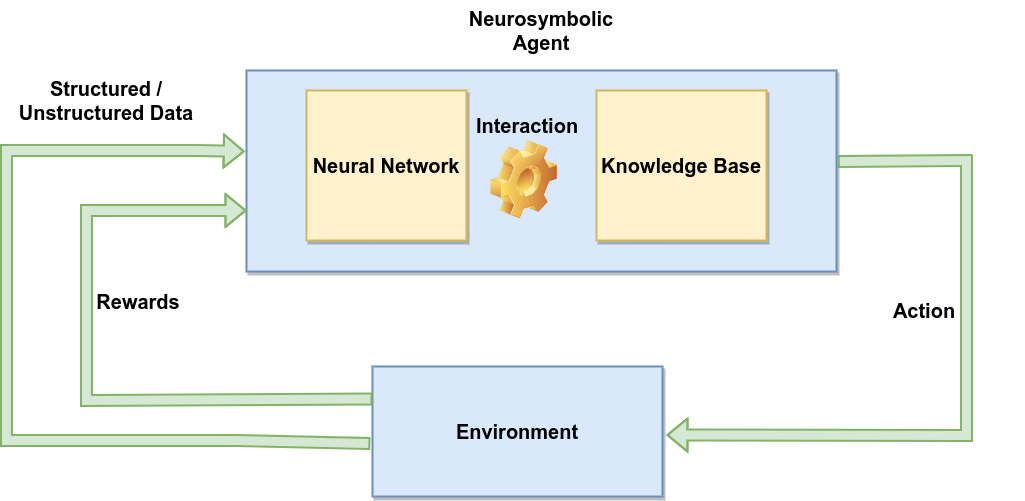}}
\caption{An Overview of Neurosymbolic RL Process}
\label{fig_1}
\end{figure}

RL has recently drawn much attention in the context of Neurosymbolic AI for policy synthesis and representation. Neurosymbolic Reinforcement Learning(Neurosymbolic RL) merge planning-style control-flow instructions with fundamental atomic actions that are learned and represented through Deep Neural Networks(DNNs). The combination of neural and symbolic approaches enables the efficient use of DRL techniques to improve the interpretability and transparency of an agent's behavior while also leveraging a high-level, symbolic representation of the policies learned by agents. By allowing the neural system to interact with the knowledge base, the reasoning ability is enhanced, and the learning ability is enhanced by interacting with the neural system. This interaction results in better generalization and transfer of knowledge, improved efficiency and robustness, and an increase in explanation and interpretability. $\hyperref[fig_1]{Fig. 1}$ illustrates the general idea of combining Neurosymbolic AI with RL to give rise to Neurosymbolic RL. Neurosymbolic agent has advantage of using both neural and symbolic counterpart so as to enhance not only its learning ability but also reasoning skills. Further advantage of such kind of model can be in the areas like reward shaping, reducing the complexity of the environment and also synthesizing the efficient symbolic policy. However this approach is on its primitive phase much of the areas are still worth exploration.

Further research is needed in Neurosymbolic RL to develop novel approaches, techniques, and their real-time applications that best fit real-world use cases such as computer networks, healthcare, IoT devices, finance, and other industrial domains. In this paper, we have analyzed notable works in Neurosymbolic RL to date. We have examined the neural and symbolic component used in each of the research work. Further we analyzed RL components used in the architecture: RL-algorithms, state space, action space, and policy module used so that we can have transparent view of the working of the model. We have classified these research works into three main categories: Learning for Reasoning RL model, Reasoning for Learning RL model, and Learning-Reasoning RL model, which are further sub-divided according to the significance or role of the model. After that we move on to provide a comprehensive summary of the Neurosymbolic RL approaches related to their specific application cases that need to be developed to meet the needs of AI. Finally, we have presented certain challenges specific to each application case to employ them in real-world scenarios.

In reviewing the history of AI surveys, no significant work has been found that specifically focused on the combination of Neurosymbolic AI with RL. Earlier works are either focused solely on Neurosymbolic AI or on the field of RL. Negligible surveys can be found on Neurosymbolic AI\cite{besold2017neural,wang2022towards} few other provides insight on recent advances\cite{yu2021survey} and application\cite{bouneffouf2022survey}.
A large number of surveys are available on RL on various aspects:
\begin{itemize}
    \item RL in general\cite{kaelbling1996reinforcement}, DRL\cite{arulkumaran2017deep,li2017deep}, Causal RL\cite{zeng2023survey}
    \item Safety and Security in RL \cite{garcia2015comprehensive, uprety2020reinforcement}
    \item Environment \cite{padakandla2021survey}
    \item Agent \cite{bucsoniu2010multi,canese2021multi,article}
    \item Application like Natural Language Processing \cite{uc2023survey,luketina2019survey}, Communication Network \cite{qian2019survey}, Robotics \cite{kober2013reinforcement}, Healthcare \cite{10.1145/3477600}, Transportation \cite{9146378}
\end{itemize}
This survey is the first of its kind and the first attempt to evaluate the combination of these two popular areas as one (Neurosymbolic RL). In this survey, we provide insights into all the relevant work done in the past under various taxonomies, along with possible opportunities to address the challenges.
 
The following is the document's structure: Section II provides an overview of milestones in the AI field from its inception to the present day. In Section III, we present an overview of Neurosymbolic AI and RL, covering relevant literature and significant research findings. Section IV is dedicated to Neurosymbolic RL, including workable architectures and requirements. In Section V, we summarize notable research in Neurosymbolic RL under various headings. In Section VI, we discuss opportunities that have emerged from Neurosymbolic RL. Section VII is devoted to the challenges of implementing proposed Neurosymbolic RL applications. We identify the obstacles and challenges that may arise. Finally, in Section VIII, we offer concluding remarks on our survey paper.

\section{Milestone in Reinforcement Learning}

    Reinforcement Learning (RL) has a rich history dating back to the 1940s when B.F. Skinner introduced the concept of operant conditioning in psychology, while Walter Pitts and Warren McCulloch\cite{mcculloch1943logical} presented a computational model based on the functioning of the human brain. Donald Hebb's Hebbian Learning Rule\cite{wang2017origin} introduced in 1949 also formed the basis for modern neural networks. In the same year, there was use of Monte Carlo method in the nuclear reactor for predicting the behaviour of neutron during second world war\cite{metropolis1949monte}.  In the late 1950s, two important concepts: Dynamic Programming\cite{bellman1957dynamic} and  Markov Decision Process(MDP)\cite{bellman1957markovian} were purposed which developed the mathematical formulation of RL, also Frank Rosenblatt developed the perceptron\cite{rosenblatt1957perceptron}, which could learn based on associationism. Later, Temporal Difference Learning (TD Learning)\cite{samuel1959some} was introduced by Arthur Samuel, which enabled agents to learn from delayed rewards and gradually update their value estimates. Grigoryevich\cite{4308320} used complex polynomial equations to statistically analyze network elements, selecting the best ones for the next layer, laying the groundwork for what would become deep learning.
    In the 1970s, the field of AI experienced a reduction in funding, leading only a few scientists to continue their work independently. Nonetheless, during this time, significant progress was achieved. Fukushima developed the Neocognitron neural network\cite{fukushima1980neocognitron}, which utilized a hierarchical multi-layer architecture to enable computers to learn visual patterns. The Neocognitron later served as a basis for the Convolutional Neural Network(CNN)\cite{lecun1989backpropagation}, introduced in 1989.

    The 1980s saw the emergence of the field of RL with the introduction of Actor-Critic Algorithms\cite{barto1983neuronlike} and Q-learning\cite{watkins1992q}. Additionally, Paul Werbos introduced the backpropagation algorithm\cite{rumelhart1986learning}, which, although not widely used at the time, raised questions in cognitive psychology regarding the role of symbolic logic in human comprehension. In the following decade, the field continued to evolve with the introduction of core algorithms such as TD-Gammon\cite{tesauro1991practical} REINFORCE\cite{williams1992simple},Experience Replay\cite{watkins1989learning} and State-Action-Reward-State-Action(SARSA)\cite{lin1992reinforcement}. A pivotal breakthrough occurred in 1999 with the invention of the Graphics Processing Unit (GPU)\footnote{\url{ https://www.computer.org/publications/tech-news/chasing-pixels/nvidias-geforce-256 }}, which enabled RL to tackle more complex environments. This was further enhanced by the parallel computing power of NVIDIA's Compute Unified Device Architecture (CUDA)\cite{lindholm2008nvidia} on GPUs.

    The 2010s proved to be a remarkable decade for RL. In 2012, the introduction of the Arcade Learning Environment (ALE)\cite{bellemare2013arcade} opened the gateway to the use of RL in gaming environments. DRL, which combines neural networks with RL to learn high-dimensional state-action value functions, was introduced, leading to breakthroughs in game playing and robotics, such as the Deep Q-Network (DQN)\cite{mnih2013playing}. Many researchers became active in modifying existing algorithms, resulting in the development of numerous new algorithms in the RL domain in 2015, such as Trust Region Policy Optimization (TRPO)\cite{schulman2015trust}, Deep Deterministic Policy Gradient (DDPG)\cite{lillicrap2015continuous}, Double DQN\cite{van2016deep}, Dueling DQN\cite{wang2016dueling} and Prioritized Experience Replay \cite{schaul2015prioritized}. Year 2016 saw the rise of algorithms like Asynchronous Advantage Actor-Critic (A3C)\cite{mnih2016asynchronous},Generative Adversarial Imitation Learning (GAIL)\cite{ho2016generative}. The introduction of OpenAI Gym\cite{brockman2016openai}, an open-source toolkit for developing and comparing reinforcement learning algorithms, opened the door for exploring RL algorithms among the RL community members. In same year of 2016, RL achieved a significant milestone in the history of AI by defeating the world champion in the game of Go\cite{silver2016mastering}.Year of 2017 also came up with new RL algorithms like Model-Agnostic Meta-Learning (MAML)\cite{finn2017model}, Distributional RL\cite{bellemare2017distributional},Proximal Policy Optimization (PPO)\cite{schulman2017proximal},Intrinsic Curiosity Module (ICM)\cite{pathak2017curiosity} and Rainbow DQN\cite{hessel2018rainbow}. RL model for the first time could learnt by playing itself and was able to beat its previous version which beat the world champion in the game of Go\cite{silver2017mastering}. RL Algorithms like Soft Actor Critic(SAC)\cite{haarnoja2018soft}, Quality Value Iteration Optimization (QT-Opt)\cite{kalashnikov2018qt} and Imapala\cite{espeholt2018impala} were introduced in year 2018. RL continued to conquer many other games against humans, such as chess and shogi\cite{silver2017masteringgames} and Starcraft II\cite{Team2019-ho}. In 3D games as well RL model gave the human level performance\cite{jaderberg2019human}. 

\begin{figure}
\centerline{\includegraphics[width=18.5pc]{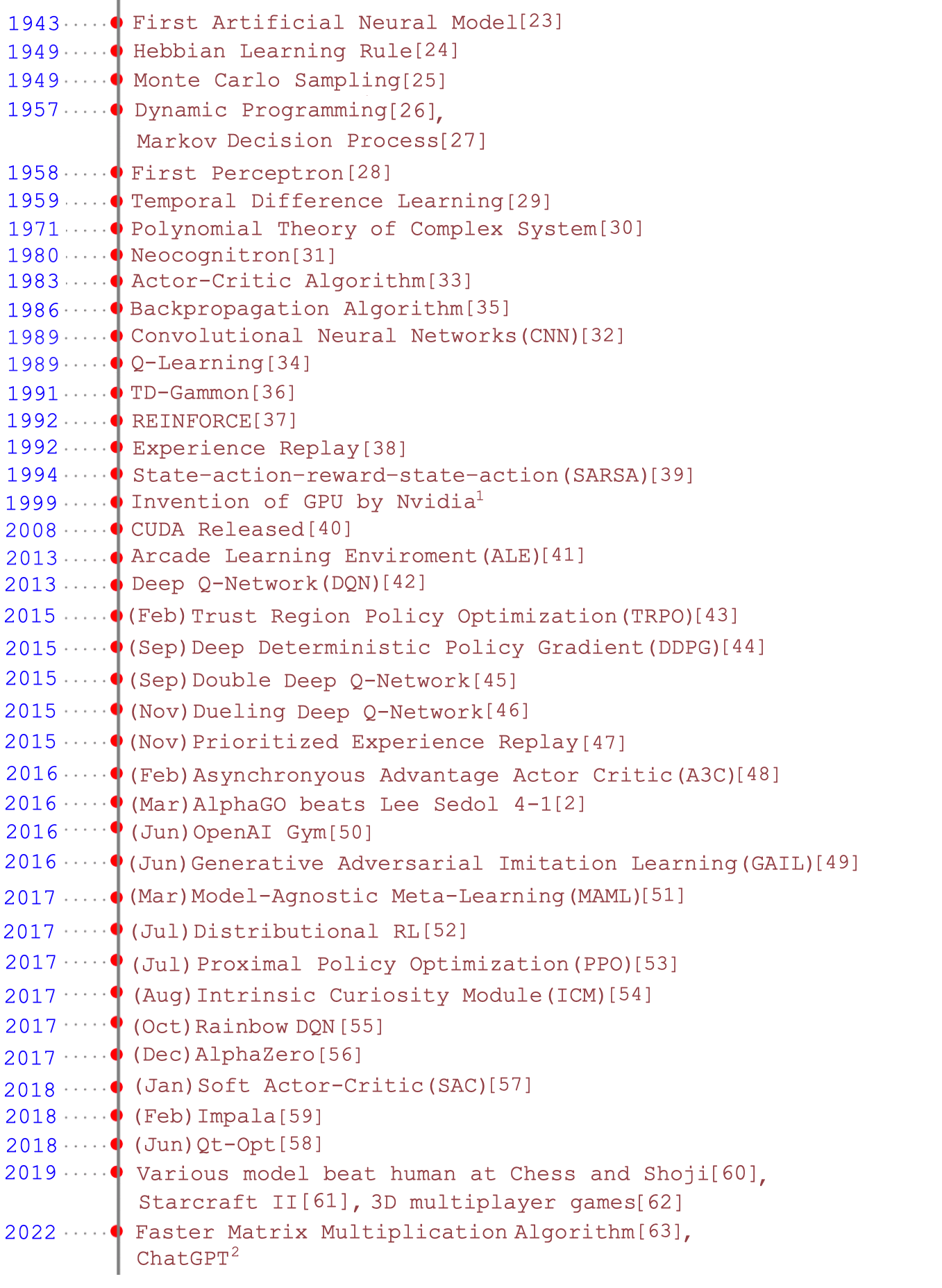}}
\caption{RL Milestone Timeline}
\label{fig_2}
\end{figure}

    As we have progressed into the 2020s, RL continues to be a dynamic field of research, with researchers exploring novel algorithms and applications, such as multi-agent RL, meta-RL, and RL for safety-critical systems. In addition, Alphazero has been successful in discovering faster multiplication algorithms\cite{fawzi2022discovering} which showed that RL can also contribute in other field also as a superhuman. Recently, in late 2022, OpenAI released ChatGPT\footnote{\url{ https://openai.com/blog/chatgpt }}, a chatbot that utilizes RL techniques to be trained and generate diverse responses to various inquiries and concerns. Reinforcement Learning from Human Feedback (RLHF)\cite{stiennon2020learning} has been introduced to enhance the alignment of the model with human values such as helpfulness, honesty, and harmlessness. RLHF aims to leverage the expertise and knowledge of humans to accelerate the learning process of an AI agent, allowing it to learn more efficiently and effectively. This involves fine-tuning using feedback data collected from humans. Other major technology companies are also in the race to develop their own innovative AI solutions. A summary of significant milestones in the history of RL is presented in $\hyperref[fig_2]{Fig.2}$.
    
\section{Background and Preliminaries}

In this section, we keep focus on the background information related to the topics Neurosymbolic AI followed by RL.

	\subsection{Neurosymbolic AI}
 The field of artificial intelligence has been centered around the goal of developing machines that can achieve human-like levels of intelligence. Two major approaches have been pursued in this effort. The first, symbolic AI, is a rule-based approach that was prevalent from the 1950s to the 1980s. The second approach is a data-based approach known as connectionist AI. While symbolic AI requires a large amount of information to be supplied, it can learn from this information on its own. The primary disadvantage of connectionist AI is its inability to explain the reasoning or logical processes behind the model, leading to these models being referred to as black boxes. Symbolic reasoning provides an explainable inference process and employs powerful declarations to represent knowledge, as well as offering benefits such as fast initial coding, explicit method control, and abstraction of knowledge\cite{mao2019neuro}. However, this approach is limited in its ability to handle vast amounts of incomplete data and to generalize from such data.
Psychologist Daniel Kahneman has distinguished between two human cognitive processes, system 1 and system 2. System 1 is fast, automatic, and unconscious, akin to deep learning, while system 2 is slow, effortful, and conscious, similar to symbolic AI\cite{kahneman2011thinking}. In the context of AI, there have been discussions of ways to combine these two approaches, as the authors of a study\cite{booch2021thinking} conclude that only a combination of both fields is likely to enable the development of human-like intelligence.

\begin{figure}
\centerline{\includegraphics[width=18.5pc]{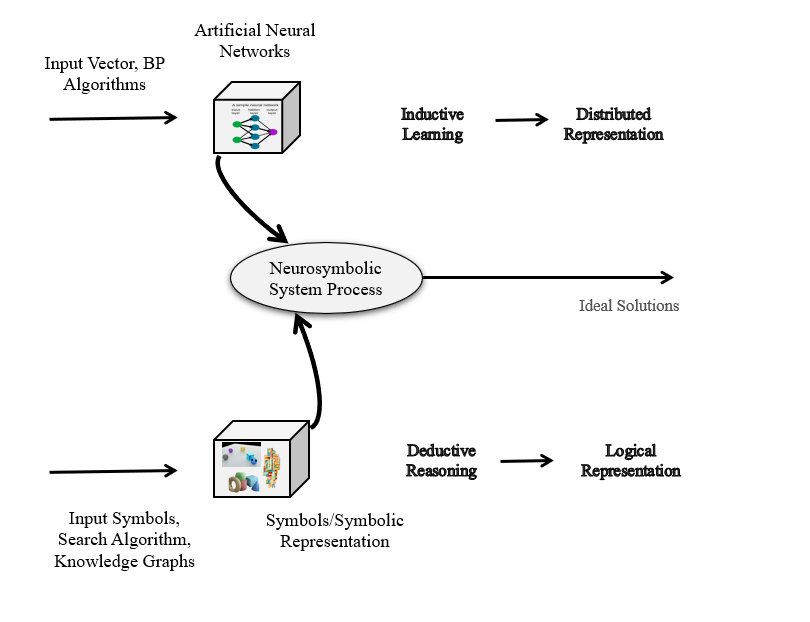}}
\caption{General Representation Neurosymbolic AI}
\label{fig_3}
\end{figure}

  Neurosymbolic AI is a subfield of AI that combines two historically prominent approaches: connectionist AI and symbolic AI. This integration enables more efficient derivation of knowledge and general concepts from data, focusing on learning from experience and reasoning about what has been learned from uncertain environments. Hybrid Neurosymbolic systems require less training data and are capable of tracking the steps required to draw conclusions and make inferences which is the reason Neurosymbolic AI has been regarded as the $3^{rd}$ wave of AI\cite{garcez2023neurosymbolic}. By combining symbolic reasoning with deep learning, ideal results can be obtained with a limited number of datasets, error correction with recoveries, and enhanced explanatory capabilities that are not possible with deep learning alone\cite{lamb2020graph}. Numerous applications necessitate both learning and reasoning abilities. On the neural aspect, models learn from data provided to them, while the symbolic aspect seeks to retain the innate explanatory power of these systems. The Neurosymbolic AI domain, as previously discussed, can be employed to develop various applications across different fields, such as medical diagnostic systems, recommender systems, and text mining\cite{sarker2021neuro}. By incorporating deep human expert knowledge into the system's design and function, Neurosymbolic AI can be leveraged to its fullest potential in creating such applications.$\hyperref[fig_3]{Fig.3}$ depicts the evaluation of the Neurosymbolic AI process within a model design that integrates neural network and symbolic artificial intelligence, harnessing the full strength of both fields in hybrid models.

\begin{figure}[h]
\centerline{\includegraphics[width=18.5pc]{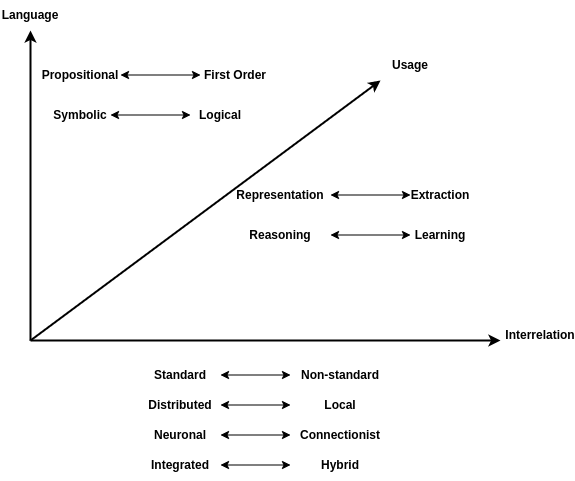}}
\caption{Classification by Sebastian and Pascal\cite{bader2005dimensions}}
\label{fig_4}
\end{figure}

Numerous researchers have provided insights into how the fields of neural and symbolic AI can be combined practically. Three noteworthy works have contributed significantly to organizing the research in Neurosymbolic systems. The first notable work was a survey paper published in 2005 by Sebastian and Pascal \cite{bader2005dimensions}. They identified three main axes of Neurosymbolic integration: Interrelation, Language, and Usage. Each of these axes was further divided into several sub-divisions. $\hyperref[fig_4]{Fig.4}$ provides a simplified visualization of the eight dimensions along with their axes. Another researcher, Henry Kautz\cite{kautz2022third}, proposed a way to classify Neurosymbolic systems into six different categories. He gave them distinctive names, which are detailed in Table I. In a recent survey\cite{yu2021survey}, Neurosymbolic systems were analyzed based on three parameters: efficiency, generalization, and interpretability. The authors proposed a novel taxonomy consisting of three different classes: learning for reasoning, reasoning for learning, and learning-reasoning. Table II provides detailed descriptions of each of these classes.

\begin{table}[htbp]
\caption{Classification by Henry Kautz\cite{kautz2022third}}
\begin{center}
\begin{tabular}{|c|c|}
\hline
Classification & Characteristic Features \\
\hline
Symbolic Neuro symbolic & \makecell[l]{ Symbolic input is converted to feature\\ vectors for the neural networks which \\ give final results in the symbolic form } \\
\hline
Symbolic[Neuro] & \makecell[l]{Neural pattern recognition subroutine \\within a symbolic problem solver} \\
\hline
Neuro $|$ Symbolic & \makecell[l]{A cascade from neural network into \\symbolic reasoner} \\
\hline
Neuro: Symbolic → Neuro &  \makecell[l]{Symbolic rules are input which are\\ compiled so that their knowledge \\end up in the neural network}\\
\hline
Neuro\_\{Symbolic\} &  \makecell[l]{Uses direct encodings of logical statements\\ into neural structures}\\
\hline
Neuro[Symbolic] & \makecell[l]{Embed symbolic reasoning inside neural \\engine to enable both superhuman and \\super  combinatorial reasoning} \\
\hline
\end{tabular}
\label{tab1}
\end{center}
\end{table}

\begin{table}[htbp]
\caption{Classification by D. Yu and et al. \cite{yu2021survey}}
\begin{center}
\begin{tabular}{|c|c|}
\hline
Classification & Characteristic Features \\
\hline
Learning for reasoning & \makecell[l]{Neural network play the role of the helper, it \\extracts the important symbols and information\\ so that the search space of the symbolic \\system narrowed down}  \\
\hline
Reasoning for learning & \makecell[l]{Symbolic system act as a helper, it provides\\ symbolic knowledge to the neural network\\ from where the final decision is made}  \\
\hline
Learning-reasoning &  \makecell[l]{Uses symbolic and neural systems as an \\alternate process. They both complement each \\other to give the final results} \\
\hline
\end{tabular}
\label{tab2}
\end{center}
\end{table}

\subsection{Reinforcement Learning}
\begin{figure*}[htbp]
\centerline{\includegraphics[width=\textwidth,height=10pc]{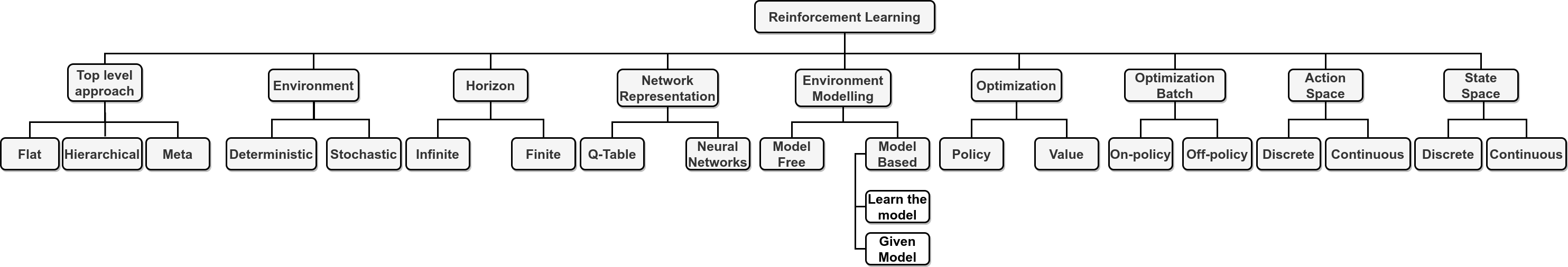}}
\caption{RL Methods and Techniques}
\label{fig_5}
\end{figure*}
The fast-learning algorithms and wide-ranging applications of RL have made it increasingly popular in academia and industry, thanks to significant technological advancements \cite{salvador2020reinforcement, balhara2022survey}. In earlier literature, RL was described as a class of problems that an agent encounters in a dynamic, unpredictable environment and solves through trial and error. Nowadays, RL is viewed as a machine learning paradigm that trains an agent to make decisions based on its immediate surroundings to optimize rewards. The training process involves a loop of interaction with the environment, including observing, receiving rewards, making decisions, and obtaining feedback signals \cite{gosavi2009reinforcement}. RL has proven its ability to solve complex real-world problems, such as natural language processing, image classification, speech recognition, and decision-making, which has improved planning and perception in various applications \cite{Wang2020}.RL is an essential component of autonomous driving cars and robots, which can perform tasks such as food preparation without human intervention or specific programming. RL-based strategies could play a crucial role in enabling fully autonomous systems in the future \cite{milani2022survey}. RL employs algorithms and methods to enable an agent to obtain optimal control in an environment, and the agents in RL can range from a game player to a stock trading bot. Another field which is similar to RL is Intrinsic Motivation(IM) but it lack feedback mechanism. Many research papers in RL have utilized IM to address complex problems in sparse reward platforms \cite{aubret2019survey,yadav2022survey}.
The interactions that occur between an agent and its environment are typically modeled as a Markov Decision Problem (MDP)\cite{Puterman1990Chapter8M}, or a Partially Observable Markov Decision Process (POMDP). An MDP is a framework used for sequential decision-making in Markovian dynamical systems, which extends the Multi-Armed Bandits (MAB) framework by allowing the system state to stochastically change based on the actions taken and their resulting outcomes. On the other hand, a POMDP is a newer version of MDP, where the system state is not directly observable. In certain cases, MDPs can be solved analytically, while in many cases, they can be solved iteratively through the use of dynamic or linear programming. When no model is present, RL methods can be employed to obtain sample trajectories and directly interact with the system\cite{qin2022reinforcement}. As the number of computing devices continues to increase rapidly, it is expected that the number of devices capable of handling complex and dynamic systems with minimal programming will grow exponentially, potentially reaching billions.  

Looking at the bigger picture, RL can be classified as a type of sample-based approach for solving MDP problems. The RL technique uses sample trajectories and the agent's interaction with the system, which can be obtained from a simulation. This approach is quite common in practical applications, where a simulation is available, and a clear transition-probability model is not required. In such scenarios, dynamic or linear programming may not be suitable, making the RL method a more practical option\cite{arulkumaran2017deep}.Main Components of Reinforcement Learning are:
\begin{itemize}
\item
Policy: 
It refers to the way an agent behaves at a given time, which is generally a mapping from perceived states to the action that needs to be taken when in those states of the environment. The goal of the policy is to maximize the expected cumulative reward received by the agent over time. There are various types of policies, including deterministic policies and stochastic policies, which define the agent's behavior in different ways.
\item
Reward Signal: 
It refers to the objective or goals of the problem and is a number delivered to the agent by the environment at each time step. The reward signal is used to train the agent to learn a behavior that maximizes the cumulative reward over time. It is a crucial component as it guides the agent to take actions that lead to achieving the desired goals.
\item
Value Function: 
It represents the expected long-term cumulative reward that an agent can obtain by following a specific policy. It estimates the value of each state or state-action pair, which allows the agent to choose the best action in each state. The value function can be expressed mathematically as the expected sum of discounted future rewards starting from a given state or state-action pair. The estimation of the value function can be done through various methods, such as Monte Carlo methods, Temporal Difference learning, or Bellman equations.
\item
Model of the Environment: 
It refers to the representation of how the environment behaves in response to the actions taken by the agent. It allows the agent to predict the next state and reward given the current state and action. The model can be either known or unknown, and the goal is to use it to optimize the agent's behavior. In cases where the model is known, dynamic programming techniques can be used to find the optimal policy. The model can be represented in different forms, including transition probabilities, state-transition diagrams, or function approximators.
\end{itemize}

Reinforcement Learning (RL) can be categorized into various types based on different parameters, such as the environment, policy, model, and others. These categories provide a framework to understand and classify different RL approaches. $\hyperref[fig_5]{Fig.5}$ summarizes the various types of RL based on these parameters, including environment type, horizon type, optimization type, and more.

\begin{figure}
\centerline{\includegraphics[width=18.5pc]{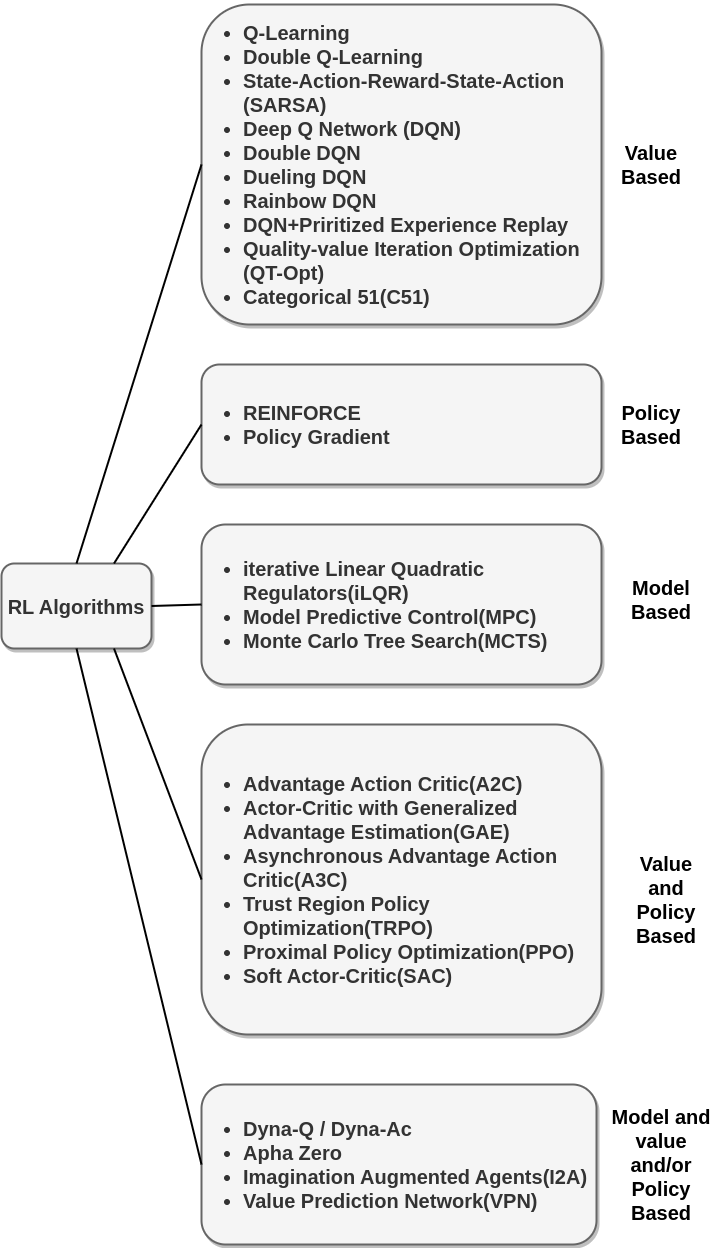}}
\caption{Overview of RL Algorithms}
\label{fig_6}
\end{figure}

The success of RL largely depends on the quality of its algorithm. Numerous RL algorithms have been developed, tailored to specific contexts, and based on various parameters such as environment type, action space type, and model type. These algorithms are continuously modified to improve their performance and expand their scope of applications\cite{zhu2021overview}. $\hyperref[fig_6]{Fig.6}$ provides a brief overview of the various types of RL algorithms that have been used to date.

\section{Neurosymbolic Reinforcement Learning}

RL, a long-standing topic in the field of AI, has faced the curse of dimensionality, but the introduction of DRL solved this problem. However, DRL has several limitations. For instance, DRL can be extremely data-inefficient. In a paper by Deepmind \cite{hessel2018rainbow}, they demonstrated that the Rainbow DQN method can achieve state-of-the-art performance in terms of both performance and data efficiency. Nevertheless, it required almost 83 hours (about 3 and a half days) of playtime in addition to the training time. Conversely, many people can achieve this level of performance in just a few minutes. Another issue with DRL is that, except for rare scenarios, domain-specific algorithms work better than DRL. In the field of robotics, Boston Dynamics\footnote{\url{ https://www.bostondynamics.com/research }} is a leading research institution that focuses mainly on classical robotics techniques such as time-varying  Linear Quadratic Regulator(LQR), Quadratic Programming(QP) solvers, and convex optimization. Another main issue with RL is the reward system, which can be easily functionalized, but the challenge arises when trying to encourage appropriate behavior while still making it learnable. Sparse rewards are problematic because they only supply rewards in the goal state, making them difficult to shape. Shaped rewards are easier to learn because they provide positive feedback even when the whole solution has not yet been figured out. However, the problem with shaped rewards is that they are biased. The agent becomes focused on maximizing the reward instead of finding the complete solution. For example, in a study on text summarization \cite{paulus2017deep}, the RL model focused on increasing the ROUGE score, which it succeeded in doing. However, it failed to achieve the actual task of generating readable summaries. In contrast, summarized text generated by the model with lower ROUGE scores was found to be more readable and efficient.  

The combination of Neurosymbolic systems and RL appears to be a solution to many of the issues identified in previous DRL methods. This approach not only adds reasoning and explaining capabilities to DRL but also provides a breakthrough in the field of RL. There are multiple ways in which the Neurosymbolic counterpart can be combined with RL, each with its own unique features. In this context, we will discuss three different approaches. 
\subsection{Learning for Reasoning RL model} 
\begin{figure}
\centerline{\includegraphics[width=18.5pc]{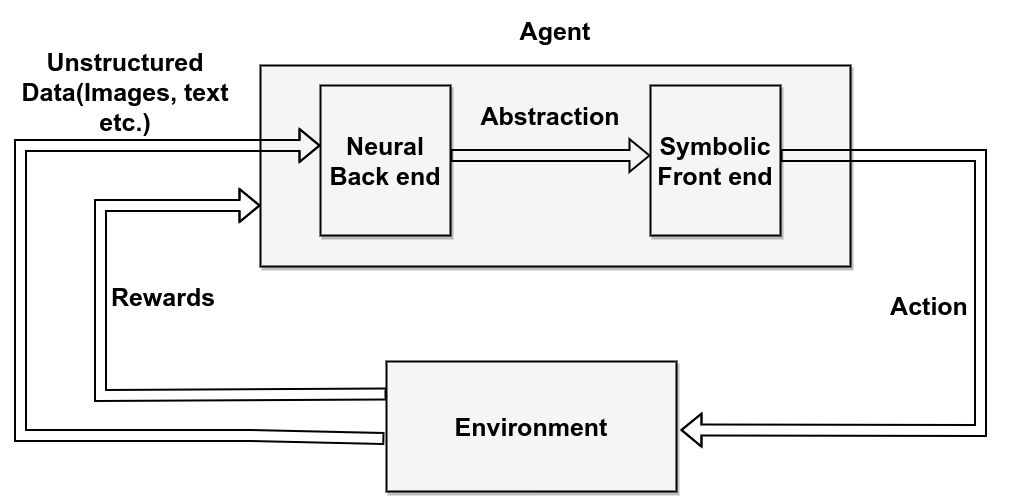}}
\caption{Learning for Reasoning RL model}
\label{fig_7}
\end{figure}

The Learning for Reasoning RL model combines a neural component with a symbolic system to improve reasoning capabilities. The neural component functions as a co-actor, while the symbolic system handles the problem of reasoning. The DNNs in the model help to reduce the symbolic space, leading to faster convergence and improved performance. In cases where the data presented to the model are unstructured, DNNs can transform them into a symbolic form that the symbolic system can utilize. Furthermore, DNNs can also distill the learning policy to the symbolic system, which enhances verifiability. Serialization characterizes the neural and symbolic counterparts in this model, as shown in $\hyperref[fig_7]{Fig.7}$.

Possible uses of the model can be:

\begin{itemize}
    \item Handling Unstructured Data: This type of model can handle unstructured data by using the neural part to transform it into a symbolic form that the symbolic system can work with. This is particularly useful when dealing with real-world data that may be unstructured or lack a clear symbolic representation.
    \item Faster Convergence and Improved Performance: By leveraging DNNs, the model can benefit from their ability to learn complex patterns and generalize from data. This lead to faster convergence during training and improved performance in terms of accuracy and efficiency.
    \item Verifiability and Interpretability: The model enhances verifiability by distilling the learned policies from the neural component to the symbolic system. The symbolic system can understand and interpret the reasoning process, making it easier to verify and understand the decisions made by the model.
\end{itemize}

\subsection{Reasoning for Learning RL model} 
\begin{figure}
\centerline{\includegraphics[width=18.5pc]{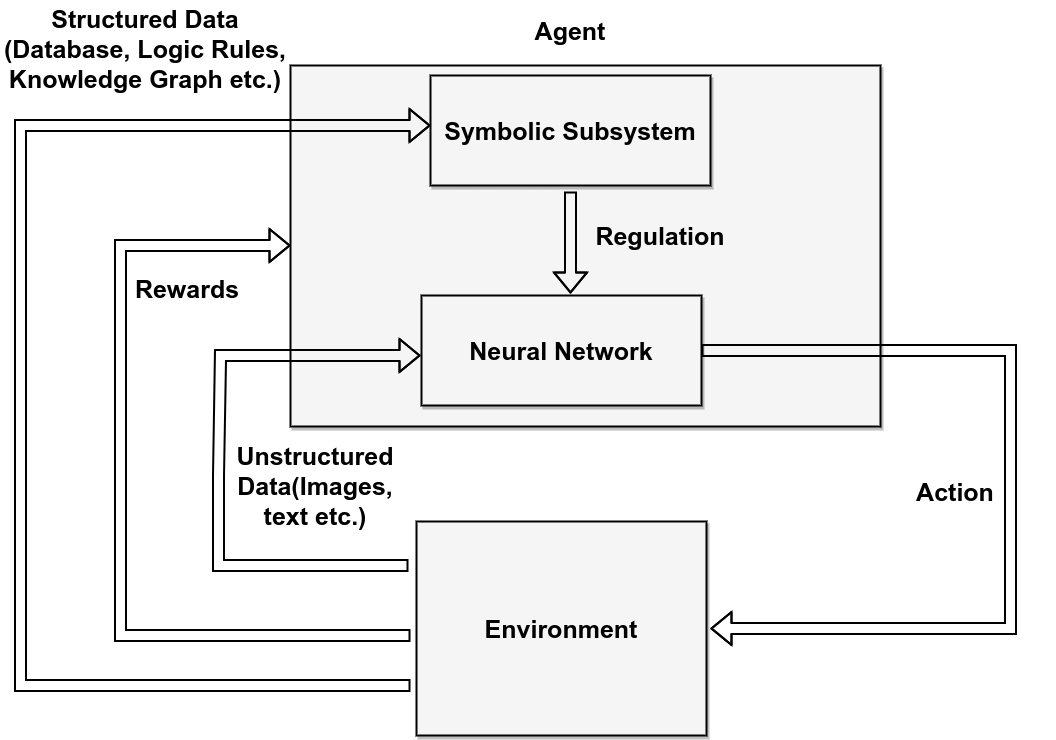}}
\caption{Reasoning for Learning RL model}
\label{fig_8}
\end{figure}

The Reasoning for Learning RL model is a different approach that utilizes symbolic models to guide the output of the neural network. By incorporating structured knowledge from the symbolic system, the performance and interpretability of the DNNs can be improved. The symbolic model can also help with reward shaping to enable faster convergence and improved performance of the DNNs agent. Additionally, the symbolic system can aid in generating the programmatic policy, making the RL model more interpretable and explainable. This type of model is characterized by parallelization, as shown in $\hyperref[fig_8]{Fig.8}$.
Application areas of the Reasoning for Learning RL model can be:
\begin{itemize}
    \item Improved Performance and Interpretability: By incorporating structured knowledge from the symbolic system, the model enhances the performance and interpretability of the neural network component. The symbolic system supplies guidance and constraints to the neural network, improving its decision-making and enabling more transparent and understandable outputs.
    \item Efficient Reward Shaping: The symbolic system in the RL model can assist in reward shaping, which involves designing reward functions that guide the learning process. By leveraging the structured knowledge from the symbolic system, reward shaping can be more effective, leading to faster convergence and improved performance of the agent.
    \item Interpretable and Explainable RL: The model's integration of a symbolic system allows for the generation of programmatic policies. This means that the RL model's decision-making process can be expressed in a human-readable and interpretable form, making it easier to understand and explain the reasoning behind the model's actions.
\end{itemize}

\subsection{Learning-Reasoning RL model} 
\begin{figure}
\centerline{\includegraphics[width=18.5pc]{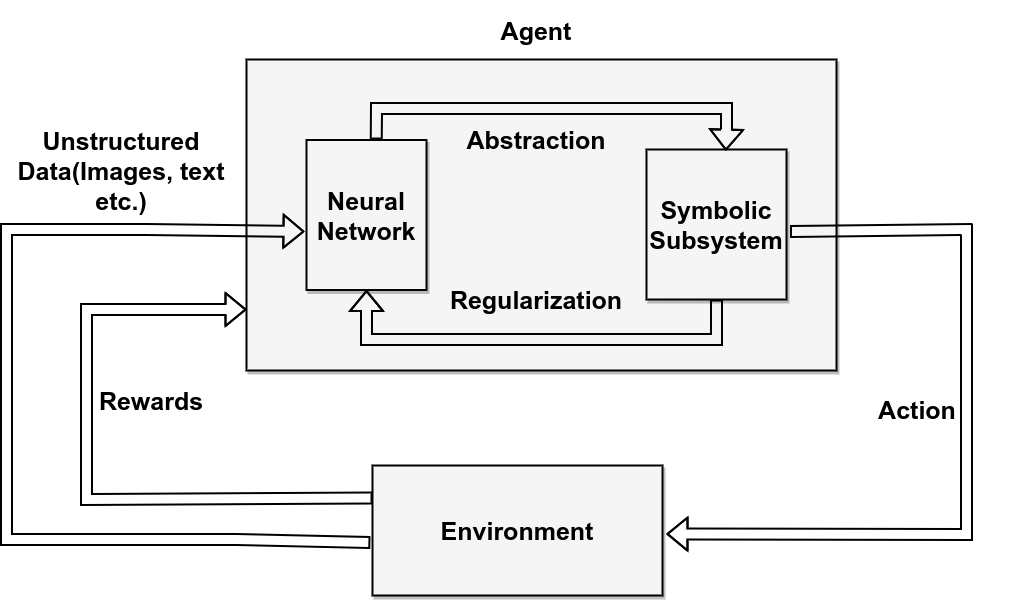}}
\caption{Learning-Reasoning RL model}
\label{fig_9}
\end{figure}

In the Learning-Reasoning RL model, the neural and symbolic components work bidirectionally, where the output of one can be the input of the other. This approach combines the benefits of both the Learning for Reasoning RL and Reasoning for Learning RL models, resulting in a balanced combination of interpretability and reasoning capacity. The symbolic part provides the structured knowledge to DNNs to enhance their interpretability and performance, while the neural component reduces the symbolic space, enabling the symbolic counterpart to achieve faster convergence. Moreover, the two parts work together which make the RL model more interpretable and explainable. This type of model is characterized by bidirectional communication, as depicted in $\hyperref[fig_9]{Fig.9}$. Some of the areas where this model can contribute are:

\begin{itemize}
    \item Enhanced Interpretability and Reasoning Capacity: By integrating the neural and symbolic components bidirectionally, the model achieves a balanced combination of interpretability and reasoning ability. The symbolic part provides structured knowledge to the DNNs, improving their interpretability, while the neural component reduces the symbolic space and enables faster convergence of the symbolic counterpart. This combination enhances the overall interpretability and reasoning capacity of the RL model.
    \item Improved Performance and Decision-Making: The bidirectional communication between the neural and symbolic components allows for an exchange of information and knowledge. Structural from the symbolic system can guide the decision-making of the neural network, leading to improved performance and more informed choices. At the same time, the neural component can transform unstructured data into a symbolic form that the symbolic system can utilize, enabling the integration of both types of information for more effective decision-making.
    \item Handling Complex and Hybrid Domains: This model is particularly well-suited for complex and hybrid domains that require a combination of symbolic reasoning and neural network learning. It can handle scenarios where structured knowledge, reasoning, and interpretation are essential, while also benefiting from the learning capacity and flexibility of neural networks.The neural component can distill the learning policy to the symbolic system, enhancing its verifiability and enabling more efficient learning. The symbolic system, in turn, provides guidance and constraints to the neural network, aiding in reward shaping and reducing the search space for better convergence.
\end{itemize}

The aforementioned three RL models are currently in their initial stages, undergoing extensive research to optimize their utilization. Table III presents a comparative summary of all three RL models.

 \begin{table*}[]
 \caption{Comparison of Neurosymbolic RL models}
\centering
\begin{tabular}{|c|l|l|l|}
\hline
Parameters             & Learning for Reasoning RL       & Reasoning for Learning RL model & Learning-Reasoning RL              \\
\hline
Neural Component     & \makecell[l]{Provide abstraction to symbolic\\ component}  &Generate the actions & Provide abstraction to symbolic component            \\
\hline
Symbolic Component      & Generate action   &\makecell[l]{Provide Regularization to the Neural \\component}   &\makecell[l]{Provide Regularization along with generating \\the actions}      \\
\hline
Connection Structure  & Serial &Parallel   &Bi-directional          \\
\hline
Interaction  & Unidirectional(Neural to symbolic)  &Unidirectional(Symbolic to Neural)  &Bi-directional     \\
\hline
Key Focus  & Improved reasoning &Improved learning  &Balance learning and reasoning   \\
\hline
Application Areas     & \makecell[l]{Transforming unstructured data\\into a symbolic representation,
\\Knowledge Graph Reasoning,\\
Verification,\\
Gaming
}  &\makecell[l]{Reward Shaping,\\
Programatic Policy Design,\\
Task Segmentation,\\
Knowledge Initialized Model
}  &      Task Segmentation        \\
\hline
\end{tabular}
\label{tab15}
\end{table*}

\section{Related Works}
Neurosymbolic Reinforcement Learning (RL) is an emerging area of AI that is currently lacking in literature. This section aims to provide an overview of the implementation stages of Neurosymbolic RL, including the state of the art, current trends, and proposed research studies. Various notable  works have been analyzed, detailing the neural and symbolic components used in Table III, and information about the RL algorithm, reward space, action space, and policy module in Table IV. These works have been classified into three main RL models: Learning for Reasoning, Reasoning for Learning, and Learning-Reasoning, and further sub-divided according to their areas of application. A summary of this classification can be found in Table V. The classification and analysis of these works provides insights into the use of Neurosymbolic RL models and their potential for further development in the future.

 \subsection{Learning for Reasoning RL model}
This particular Neurosymbolic RL model involves the use of a neural network as an auxiliary tool to extract crucial symbols and information, which helps to reduce the search space of the symbolic system. This results in a faster problem-solving process, making it especially useful for problems that require reasoning. This architecture has been used for the following primary goals:

\subsubsection{Transforming unstructured data into a symbolic representation}
Symbolic systems, which rely on logical rules and representations of symbolic data, are often limited in their ability to process unstructured data such as images, videos, and natural language text. Most of the real world data are inherently present in the unstructured form so there must be a model to transform them to the symbolic form before being processed by the symbolic models.  DNNs have been shown to be very effective in processing and generating such unstructured data and can be used to generate structured data that can be used as input to a symbolic system.
 
 Deep Symbolic Reinforcement Learning (DSRL)\cite{garnelo2016towards} consists of two main components: a deep neural network that learns a low-level continuous representation of the state space and map it to low-dimensional symbolic space, and a symbolic model that distills the learned policy into a more interpretable form by mapping symbolic representation to action. The authors use this framework to learn policies for a range of environments, and demonstrate that their system outperforms traditional reinforcement learning algorithms in terms of interpretability, generalization, and efficiency.Symbolic Reinforcement Learning with Common Sense (SRL+CS)\cite{garcez2018towards} a novel extension of DSRL where the authors create a meaningful symbolic representation of the world using sub-states before applying learning and decision-making algorithms. They have two modification in Q-value function: restricting the updates to specific sub-state and assigning importance on the basis of distance of the objects. This approach provides better generalization and explainability. Neural Symbolic Reinforcement Learning (NSRL)\cite{ma2021learning}, includes a reasoning module based on neural attention networks, which performs relational reasoning on symbolic states and induces the RL policy, enabling end-to-end learning with prior symbolic knowledge. It can extract the logical rules selected by the attention modules instead of storing all the rules, saving memory budget and improving scalability.Deep Symbolic Policy\cite{landajuela2021discovering}, uses an autoregressive recurrent neural network to generate symbolic policies, which are optimized using a risk-seeking policy gradient. To scale to environments with multi-dimensional action spaces, the authors propose an "anchoring" algorithm that distills pre-trained neural network-based policies into fully symbolic policies. The authors also introduce two novel methods to improve exploration in DRL-based combinatorial optimization which are hierarchical entropy regularizer and a soft length prior.

Detect, Understand, Act (DUA) \cite{mitchener2022detect}, composed of three components: Detect, which consists of a traditional computer vision object detector and tracker, Understand, which provides an answer set programming (ASP) paradigm for symbolically implementing a meta-policy over options, and Act, which houses a set of options that are high-level actions enacted by pre-trained DRL policies. The paper evaluates the DUA framework on the Animal-AI (AAI) competition testbed and achieves state-of-the-art results in multiple categories. It is modular approach, allowing for straightforward generalization and transfer to other complex tasks. Another study, Symbolic Options for Reinforcement Learning (SORL)\cite{jin2022creativity}, proposes a method for automatically discovering and learning symbolic options, which are higher-level actions with specified preconditions and postconditions, to assist deep reinforcement learning (DRL) agents in complex environments. It was successful in mitigating the problem of sparse and delayed reward along with improving efficiency.Neurosymbolic Logic Neural Network (LNN) for RL algorithm\cite{kimura2021neuro}, supplies fast convergence and interpretability for RL policies in text-based interaction games by extracting first-order logical facts from text observation using semantic parser(ConceptnNet) and history, then trains the symbolic rules with logical functions in the neural networks.

\subsubsection{Knowledge Graph Reasoning}
Knowledge Graph (KG) reasoning is the task of inferring new information from a given KG, which consists of a set of entities and their relationships. KG reasoning is important for a wide range of applications, including question answering, information retrieval, and recommender systems.

DeepPath\cite{xiong2017deeppath}, uses a policy-based agent with continuous states based on knowledge graph embeddings to sample the most promising relation and extend the multi-hop relational path. The authors demonstrate that their method outperforms path-ranking based algorithms and knowledge graph embedding methods on two standard reasoning tasks on Freebase and Never-Ending Language Learning datasets. Meandering In Networks of Entities to Reach Verisimilar Answers (MINERVA)\cite{das2017go}outperforms DeepPath, as Deeppath  cannot be applied to query answering tasks where the second entity is unknown.It uses neural reinforcement learning to learn how to navigate the knowledge graph conditioned on the input query to find predictive paths.

  \subsubsection{Verification} The process of verification aims to determine if a model meets a particular desired property, and can play a key role in enhancing quality and safety. In the context of reinforcement learning (RL), it is important to verify a model's convergence, correctness, and robustness to ensure it functions effectively.
  
Verifiability via Iterative Policy ExtRaction(VIPER)\cite{bastani2018verifiable},is a modification of Q-DAGGER algorithm used for verifying the correctness of deep reinforcement learning policies. The approach involves extracting a small, interpretable model from a deep neural network policy, which can then be verified using existing verification techniques. The extracted model approximates the original policy well and can be used to analyze the policy's convergence, correctness, and robustness.Another model, Reinforcement Learning with Verified Exploration (REVEL)\cite{anderson2020neurosymbolic}, incorporates a differentiable symbolic planner that generates a set of safe exploration actions, which the RL agent executes to find optimal policies while avoiding potentially unsafe states. The proposed framework is formally verified using the Coq proof assistant, which ensures that the system is free from runtime errors and satisfies desired safety properties.

\subsubsection{Gaming}Neurosymbolic RL is a promising approach in the field of game playing, where the goal is to develop agents that can learn to play games at a human-like level or beyond. It involves combining neural networks and symbolic systems to develop agents that can learn the rules of the game and develop strategies to play the game effectively. By combining symbolic reasoning with deep learning, Neurosymbolic RL models can provide explanations for the decisions made by the agent, making it easier for humans to understand and evaluate the agent's performance.AlphaGo Zero\cite{silver2017mastering}, achieves superhuman performance in the game of Go without using human data or domain knowledge. The system is based on a combination of deep neural networks and Monte Carlo tree search, with the neural networks trained through a reinforcement learning process using self-play. It beats the older AlphaGo version in game Go with score of 100-0.

\begin{table}
\caption{Neural and symbolic components in related works}
\begin{center}
\begin{tabular}{|c|c|c|}
\hline
      Research  & NN    & Knowledge Base                \\
\hline
\cite{garnelo2016towards}                             & CNN            & First Order Logic           \\
\hline
\cite{garcez2018towards}                             & CNN            & First Order Logic            \\
\hline
\cite{ma2021learning}                            & Transformer    & First Order Logic          \\
\hline
\cite{landajuela2021discovering}                        & RNN            & Decision Tree       \\
\hline
\cite{mitchener2022detect}      & DNN            & Answer Set Programming  \\
\hline
\cite{jin2022creativity}                       & DNN            & Propositional Logic           \\
\hline
\cite{kimura2021neuro}                                    & LNN            & First Order Logic   \\
\hline
\cite{xiong2017deeppath}                              & DNN            & Knowledge Graph          \\
\hline
\cite{das2017go}                              & LSTM           & Knowledge Graph    \\
\hline
\cite{bastani2018verifiable}          & DNN            & Decision Tree      \\
\hline
\cite{anderson2020neurosymbolic}                        & DNN            & Symbolic Policies               \\
\hline
\cite{silver2017mastering}                        & DNN            & Decision Tree                  \\
\hline

\cite{velasquez2021dynamic}                             & CNN            & Finite Trace Linear Temporal Logic        \\
\hline
\cite{velasquez2022multi}                             & CNN            & Finite Trace Linear Temporal Logic          \\
\hline
\cite{kazemi2022translating}                        & DNN            & Omega Regular Language           \\
\hline
\cite{verma2019imitation}            & DNN            & Programmatic Policy            \\
\hline
\cite{verma2018programmatically}          & DNN            & Programmatic Policy          \\
\hline
\cite{inala2020synthesizing}  &NN     &State Machines \\
\hline
\cite{trivedi2021learning}                             & RNN            & Programmatic Policy       \\
\hline
\cite{hasanbeig2021deepsynth}                                     & DNN            & Deterministic Finite Automaton                \\
\hline
\cite{silva2021encoding}                                     & DNN            & Propositional Logic       \\
\hline
\cite{lyu2019sdrl}                             & CNN            & First Order Logic       \\
\hline
\end{tabular}
\label{tab3}
\end{center}
\end{table}

 \subsection{Reasoning for Learning RL model}
In this type of Neurosymbolic RL model, symbolic system acts as a helper, it provides symbolic knowledge to the neural network from where the final decision is made. This approach is particularly useful in complex applications, such as robotics, where the environment is often uncertain and dynamic, and where the use of symbolic knowledge can facilitate high-level reasoning and decision-making. It has been used in the following application area:
 
\subsubsection{Reward Shaping}In the field of RL, one of the primary challenges is dealing with sparse rewards. One solution to this problem is reward shaping, which involves incorporating domain knowledge. Rather than relying on a single, final reward, intermediate rewards are provided to the agent for exhibiting desirable behavior. This encourages the agent to take effective actions early on in the learning process, leading to faster convergence. 

Monte Carlo Tree Search with Automaton-Guided Reward Shaping (MCTS-A)\cite{velasquez2021dynamic}, helps to improve the learning process and final performance of the agent in domains with sparse rewards. The method introduces an automaton that guides the reward shaping process, allowing for a dynamic and flexible approach. The automaton's states correspond to the different learning phases of the agent, and each state has its own shaping rules that change over time. Transfer learning between the two different environments with the same objective is also eased with this approach. Authors in\cite{velasquez2022multi}, extended the work by introducing Multiagent Tree Search Algorithm with reward shaping(MATS-A) so that it can be applied to multi-agent scenario and can handle both stochastic and deterministic transition in Multi-agent Non-Markovian Reward Decision Process. They prove that sharing the same search tree and DFA objective can be used to develop competitive and cooperative behavior among the agents, within and across the team.Research work \cite{kazemi2022translating} first converts an omega-regular specification into a Buchi automaton. It is then used to construct an average reward objective which can then be optimized by standard RL algorithms.The authors prove that the learned policy converges to the optimal policy and demonstrate the effectiveness of the method.

\subsubsection{Programmatic Policy Design}
A programmatic policy refers to a decision-making algorithm that governs the behavior of an agent that can make decisions. The programmatic policy takes inputs from the environment and computes a set of actions that the agent should take in response.The design of programmatic policies can vary based on the complexity of the task and available data, including decision trees, state machines, and programs.

Imitation-Projected Programmatic Reinforcement Learning (PROPEL)\cite{verma2019imitation}, proposes a new approach for combining imitation learning with reinforcement learning, called imitation-projected programmatic reinforcement learning (IP-PRL). The approach uses a programmatic policy to encode a priori knowledge about the task and trains an agent through a combination of imitation learning and reinforcement learning. The agent first learns to imitate an expert's behavior through supervised learning, and then the agent's policy is updated through reinforcement learning while being constrained to stay close to the expert's policy. IP-PRL outperforms both pure imitation learning and pure reinforcement learning in terms of sample efficiency and final performance. Another work Programmatically Interpretable Reinforcement Learning (PIRL)\cite{verma2018programmatically}, uses Neurally Directed Program Synthesis (NDPS) algorithm to generate interpretable neural policies which can be verified through the symbolic approach. It first learn a neural policy network using deep reinforcement learning and then performing a local search over programmatic policies that seeks to minimize the distance from this neural oracle. \cite{inala2020synthesizing} proposes a novel approach for synthesizing policies for automated decision-making systems that can generalize to new situations. The authors use inductive programming techniques, specifically "program synthesis by example", to generate policies that satisfy a set of example-based specifications. Framework\cite{trivedi2021learning} synthesize programmatic policies that are more interpretable and generalizable than neural network policies produced by deep reinforcement learning methods. It uses a program representation and only requires minimal supervision compared to prior programmatic reinforcement learning and program synthesis works. It learns a program embedding space that parameterizes diverse behaviors in an unsupervised manner and then searches over this space to find a program that maximizes the return for a given task.

\subsubsection{Task Segmentation}
Main goal or task is broken down in to the smaller task with their own set of rewards so that the task become more generalizable and reasonable. DeepSynth\cite{hasanbeig2021deepsynth},  uses automata synthesis to automatically segment a task. A task was broken down into smaller subgoals, each with its own reward. The proposed approach learns a model of an automaton that represents the state machine of a task and uses it to segment the task into subgoals. The learned automaton is then used to guide the agent in finding the optimal policy for the task. 

\subsubsection{Knowledge Initialized Model}
Researches has found that the model give higher convergence rate, reasoning ability if initialized with knowledge base.Before starting the learning process, the knowledge base of the agent is initialized with some prior information instead of starting from zero. Propositional Logic Nets (PROLONETS)\cite{silva2021encoding},enables warm start of learning process by efficient initialization of RL agents using human-specified policies, without requiring an Imitation Learning (IL) phase. This approach helps RL agents to navigate complex environments that pose challenges to randomly initialized models, and allows for greater exploration. It outperforms baseline RL approaches such as IL and knowledge-based techniques. 

 \subsection{Learning-Reasoning RL model}
This Neurosymbolic RL model uses symbolic and neural systems as an alternate process. They both complement each other by performing abstraction and regularization to give the final results.Symbolic Deep Reinforcement Learning (SDRL) \cite{lyu2019sdrl}, uses planner-controller-meta-controller architecture where planner uses prior symbolic knowledge for long term planning, controller uses DRL algorithms for intrinsic rewards and meta-controller evaluate training performance of controller based on extrinsic rewards along with proposing new intrinsic goals to the planner.

\begin{table*}
\caption{RL components of the related works}
\centering
\begin{tabular}{|c|c|c|c|c|}
\hline
       Research      & RL-Algorithm       & State Space & Action Space & Policy Module               \\
\hline
\cite{garnelo2016towards}      & Q-Learning  &Multi-dimensional vector & Multi-dimensional vector & Tabular Q Learning            \\
\hline
\cite{garcez2018towards}       & Q-Learning   &Multi-dimensional vector   &Multi-dimensional vector   & Q-Table    \\
\hline
\cite{ma2021learning}  & Double DQN &Set of Predicates   &Set of Predicates    &Multi-layer Perceptron           \\
\hline
\cite{landajuela2021discovering}  & Policy Gradients  &Multi-dimensional vector  &Multi-dimensional vector   &RNN   \\
\hline
\cite{mitchener2022detect}  & PPO &Multi-dimensional vector  &Discrete action space    &NN\\
\hline
\cite{jin2022creativity}     & Double Q-Learning  &High level dimension sapce  &5-dimensional vector    &Option Set             \\
\hline
\cite{kimura2021neuro}    & DQN    &Multi-dimensional vector &Discrete set of 10 different actions   & LNN \\
\hline
\cite{xiong2017deeppath}     & Policy Gradients       &Entities in Knowledge Graph  &Relations in Knowledge Graph    &Fully connected NN  \\
\hline
\cite{das2017go}  & REINFORCE              &Entities in Knowledge Graph &Relations in Knowledge Graph & LSTM\\
\hline
\cite{bastani2018verifiable}  &   VIPER  &Multi-dimensional vector   &Leaf nodes in Decision Tree  &Decision Tree        \\
\hline
\cite{anderson2020neurosymbolic}      & Policy Gradients &Real vector space  &Real vector space   &NN \\
\hline
\cite{silver2017mastering}      & Policy Gradients    &Muti-dimensional vector &Mutli-dimensional vector & DNN         \\
\hline
\cite{velasquez2021dynamic}       & MCTS-A   &Nodes of Automata  &Transitions of Automata &CNN      \\
\hline
\cite{velasquez2022multi}        & MATS-A  &Nodes of Automata   &Transitions of Automata &CNN      \\
\hline
\cite{kazemi2022translating}     & Differential Q-Learning &Nodes in GFM automaton &Relation in GFM automaton    &DNN \\
\hline
\cite{verma2019imitation} & Policy Gradients    &Continuous &Continuous & Programatic and Neural       \\
\hline
\cite{verma2018programmatically}  &   NDPS  &Unconstrained Policy Space &Continuous  &    Programatic and Deterministic     \\
\hline
\cite{inala2020synthesizing}  & Gradient Based Optimization &Continuous  &Continuous &State Machine \\
\hline
\cite{trivedi2021learning}      & REINFORCE        &Program Embedding Space  &Program Execution Trace    &RNN \\
\hline
\cite{hasanbeig2021deepsynth}         & DQN      &Multi-dimensional vector  &  Multi-dimensional vector &  DNN      \\
\hline
\cite{silva2021encoding}    & PPO    &193D and 37D     & 44D and 10D   & Decision Tree     \\
\hline
\cite{lyu2019sdrl}      & Double Q-Learning        &High dimensional  &Set of Primitive Action    &DNN \\
\hline
\end{tabular}
\label{tab4}
\end{table*}

\begin{table*}[htbp]
\caption{Summary of classification of related researches}
\begin{center}
\begin{tabular}{|c|l|l|}
\hline
\multicolumn{1}{|l|}{RL model}           & Areas of Application                             & Related Researches                                                                                       \\ \hline
\multirow{4}{*}{Learning for Reasoning}  & Transforming unstructured data into a symbolic representation                    & \cite{garnelo2016towards}\cite{garcez2018towards}\cite{ma2021learning}\cite{landajuela2021discovering}\cite{mitchener2022detect}\cite{jin2022creativity}\cite{kimura2021neuro}  \\ \cline{2-3} 
                                         & Knowledge Graph Reasoning                                     & \cite{xiong2017deeppath}\cite{das2017go}                                                     \\ \cline{2-3} 
                                         & Verification                                     & \cite{bastani2018verifiable}\cite{anderson2020neurosymbolic}                                                     \\ \cline{2-3} 
                                         & Gaming &  \cite{silver2017mastering}                          \\ \hline
\multirow{4}{*}{Reasoning for Learning}  & Reward Shaping                                   & \cite{velasquez2021dynamic}\cite{velasquez2022multi}  \cite{kazemi2022translating}                                                   \\ \cline{2-3} 
                                         & Programatic Policy Design                              & \cite{verma2019imitation}\cite{verma2018programmatically}\cite{inala2020synthesizing}\cite{trivedi2021learning} \\ \cline{2-3} 
                                         & Task Segmentation                                   & \cite{hasanbeig2021deepsynth}                                                     \\ \cline{2-3} 
                                         & Knowledge Initialized Model                                   & \cite{silva2021encoding}                          \\                            
                                         \hline
\multicolumn{1}{|l|}{Learning-Reasoning} & Task Segmentation                                & \cite{lyu2019sdrl}                                                                               \\ \hline
\end{tabular}
\label{tb5}
\end{center}
\end{table*}

\section{Opportunities}
    Real-world applications strive to minimize errors that can arise from risky exploration and exploitation, whereas Neurosymbolic RL methods employ a trial-and-error mechanism. However, to reconcile this contradiction between Neurosymbolic RL and real-world applications, a viable approach is to create an authentic simulator using real data and domain knowledge of the model dynamics. Subsequently, objectives can be designed for the agent, and the policy network can be trained in the simulator. Finally, the trained policy can be deployed in the real world with further enhancements. Though Neurosymbolic RL is in its early stage but it has started contributing to other RL areas as well. Causal Reinforcement Learning\cite{zhu2019causal} has been able to produce the significant result since its collaboration with Neurosymbolic RL model. In this section, we examine the opportunities of Neurosymbolic RL methods in various fields.
   
    \subsection{Robotics and Control}
    Building autonomous embodied robotic systems requires designing suitable policies that ensure the system operates within reasonable mechanical constraints while maintaining safety and data efficiency. RL has been growing in robotics from very old time\cite{kober2013reinforcement,deisenroth2013survey}. The symbolic method has been introduced in robotic motion planning and control in 2007\cite{belta2007symbolic} to address these concerns.  It is clear that decision-making is a crucial aspect of robotics control, and there have been various approaches to address this challenge. One notable technique is the Neurosymbolic Program Search (NSPS)\cite{sun2021neuro}, which produces interpretable and robust Neurosymbolic programs for autonomous driving design. Another approach is the decomposition of decision-making into two levels: what to do and how to do it\cite{silver2022learning}. This method utilizes Neurosymbolic skills and has been shown to be effective in various robotics tasks. There have also been efforts to construct robotic platforms for building manipulation environments, such as the open-source platform CausalWorld\cite{ahmed2020causalworld}. Overall, these methods and platforms have been shown to improve policy learning and performance in robotics tasks.
    
    \subsection{Gaming RL}
    Games are considered as suitable benchmarks with clear-cut rules and boundaries in the RL community. Over the past few years, gaming AI has exhibited extraordinary decision-making abilities, surpassing human-level performance in various decision-making games, including card games\cite{zha2021douzero}, board games\cite{silver2016mastering}, and video games\cite{liu2021introduction}. Neurosymbolic RL has primarily been utilized in board games and video games, yielding state-of-the-art outcomes. It is expected that these models will also outperform others in card games. However, open-ended games such as Minecraft and XLand remain unexplored.
    
    \subsection{Intelligent Question Answering}
    Neurosymbolic RL has emerged as a powerful tool in natural language processing for intelligent question answering, which involves deducing the answer to a given question based on the surrounding context, often comprising both text and images. Although various studies in the context of Neurosymbolic RL have focused on knowledge graph reasoning\cite{xiong2017deeppath, das2017go} for this task, combining both text and images remains a relatively unexplored research area. As such, there is considerable potential for future research to investigate and advance this topic.
    
    \subsection{Safe Reinforcement Learning}
    Reinforcement learning (RL) has become popular due to its ability to learn from experience and make decisions in complex environments. However, in safety-critical settings such as autonomous driving, robotics, or medical diagnosis, the failure of the system can result in severe consequences, including loss of property or human lives. Ensuring the safety of RL agents is crucial in such settings. Neurosymbolic RL has been used for the verification of the RL\cite{bastani2018verifiable,anderson2020neurosymbolic} and it has given some significant result. But, it is still in its infancy period, and there are many opportunities for safely exploring the RL.

    \subsection{Optimizing Parameters of RL}
    The RL framework consists of several components, including the environment, the agent, the policy, and the reward function. Neurosymbolic RL has been applied to different components of the RL framework, combining symbolic reasoning with neural networks to solve complex RL problems. It has been successful in addressing the issue of sparse rewards by formulating reward functions that provide more informative feedback to the agent \cite{velasquez2021dynamic,velasquez2022multi,kazemi2022translating}. It has also been used to learn programmatic policies that are more generalizable and flexible to different environments \cite{verma2019imitation,verma2018programmatically,inala2020synthesizing,trivedi2021learning}. Additionally, Neurosymbolic RL has been effective in reducing the symbolic space, resulting in more efficient representations of the policy, and improving the agent's performance \cite{garnelo2016towards,garcez2018towards,ma2021learning,landajuela2021discovering,mitchener2022detect,jin2022creativity,kimura2021neuro}. 

\section{Challenges}
Neurosymbolic RL addresses a variety of issues that were previously challenges for DRL and has opened up new opportunities for researchers to develop novel methodologies. In this section, we outline a list of problems that are still prevalent with Neurosymbolic RL, including some that are specific to DRL and others that are more general research gaps.

\subsection{Automated Generation of Symbolic Knowledge}
Neurosymbolic RL relies on an environment where the agent can interact and receive rewards. Typically, these environments are represented by symbolic knowledge, as explained in the previous section. Symbolic knowledge encompasses both logic rules and knowledge graphs. While research into the automatic construction of non-logical symbolic part like knowledge graphs is relatively mature\cite{LiuQiao2016KnowledgeGC,martinez2018openie,ji2021survey}, the automatic learning of logic rules from data remains an underexplored area. Typically, domain experts manually construct the logic which is a time-consuming, laborious, and non-scalable process. Additionally, achieving end-to-end learning for rules that describe prior knowledge from data is a challenge for Neurosymbolic systems. Moreover, the inclusion of intricate logic, probabilistic relations, or diverse data sources adds further complexity to the problem. We contend that greater attention should be given to the comprehensive and automatic discovery of symbolic knowledge, not only from increasingly vast data sets but also from networks with rapidly expanding dimensionality.

\subsection{Verification and Validation}
Neurosymbolic RL models have gained popularity across multiple industries, with their size increasing rapidly to enable deployment in larger scenarios. These models have achieved state-of-the-art results and have provided a degree of reasoning and explainability. However, due to the relative novelty of this field, there is a lack of validation and verification methods for these models, which need to be addressed. For instance, despite AI surpassing humans in the game of Go in 2016, recent adversarial attacks on the models have exposed their weaknesses and led to humans defeating similar AI models in the game which have otherwise dominated grandmasters\footnote{\url{ https://arstechnica.com/information-technology/2023/02/man-beats-
machine-at-go-in-human-victory-over-ai/ }, accessed: 23/06/2023}. This highlights the significant gap in the verification field, which requires extensive work to ensure that Neurosymbolic RL models are thoroughly validated and can be deployed without any flaws. Some work\cite{bastani2018verifiable,anderson2020neurosymbolic,karimi2020formalizing} has been initiated in this in this direction but prior work \cite{barnat2010divine,barrett2010smt} also need expansion so that they can be applied to Neurosymbolic RL domain.

\subsection{Neurosymbolic RL Algorithms}
The combination of neural, symbolic, reinforcement learning allows for a more comprehensive approach to problem-solving, as it enables the system to work with both numerical and symbolic data for RL. This provides a more powerful and flexible framework for learning, allowing for the integration of different types of knowledge and reasoning techniques for the agent.However, in order to effectively combine these fields, new learning algorithms need to be designed that can take advantage of the strengths of both symbolic and neural learning so as to be implemented in RL. The traditional reinforcement learning algorithms may not be accurate enough for Neurosymbolic learning, as they do not account for the complexities and nuances of symbolic reasoning.Therefore, new algorithms need to be optimized to work under the union of two sets of knowledge, leveraging the strengths of both neural and symbolic learning\cite{chaudhuri2021neurosymbolic}. By doing so, researchers can develop more accurate and efficient learning algorithms that can be applied to a wide range of problems in fields such as natural language processing, robotics, and healthcare, among others.

\subsection{Balancing Reasoning and Learning in RL}
Neurosymbolic RL requires training the neural components using meaningful symbolic constraints and allowing the symbolic components to evolve with high-quality data-driven rules. However, transitioning between neural and symbolic components can lead to a loss of learning or reasoning power, which presents scalability challenges for the field. One crucial issue in Neurosymbolic RL is how to align symbolic specifications with representations learned using neural methods, known as the symbol grounding problem\cite{harnad1990symbol,mooney2008learning}. This challenge is well-known in AI, but it is particularly complicated in Neurosymbolic RL, where symbolic and neural components can be interwoven in intricate ways.

\section{Conclusion}
In recent years, there has been a remarkable growth in the field of Neurosymbolic Reinforcement Learning (RL). This survey provides a comprehensive overview of Neurosymbolic RL, which can be classified into three RL models: Learning for Reasoning, Reasoning for Learning, and Learning-Reasoning. We have examined each category's core area of application and conducted an in-depth analysis of its various components, including neural, symbolic, and RL. Additionally, we have highlighted future opportunities for exploration and the potential challenges that might arise. Our hope is that this survey will inspire the AI community to delve deeper into this area and explore its possibilities.

\bibliographystyle{IEEEtran}
\bibliography{ref.bib}

\begin{IEEEbiography}[{\includegraphics[width=1in,height=1.25in,,clip,keepaspectratio]{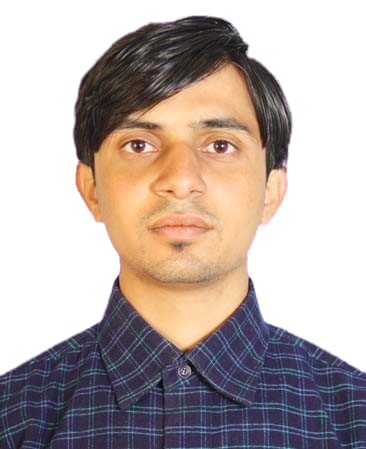}}]{Kamal Acharya}{\space}(Graduate Student Member, IEEE) received his Engineering degree in Electronics and Communication Engineering from Tribhuvan University, Kathmandu, Nepal in 2011 and Masters degree in Information System Engineering from Purbanchal University, Kathmandu, Nepal in 2019. Currently, he is pursuing PhD. in Information Systems from University of Maryland, Baltimore County (UMBC), Baltimore, MD.

He has been involved in teaching profession for about 7 years in the various universities of Nepal, Tribhuvan University and Purbanchal Univesity were among few of them. He is mainly associated with the courses like programming(C,C++,Python), Computer Networks and Computer Architecture. He is working as Graduate Research Assistant in UMBC. He is also serving as an reviewer for IEEE Transactions on Artificial Intelligence (TAI) and IEEE Transactions on Intelligent Transportation Systems. His preferred areas of research are Natural Language Processing(NLP), Deep Learning and Reinforcemnt Learning.
\end{IEEEbiography}

\begin{IEEEbiography}[{\includegraphics[width=1in,height=1.25in,,clip,keepaspectratio]{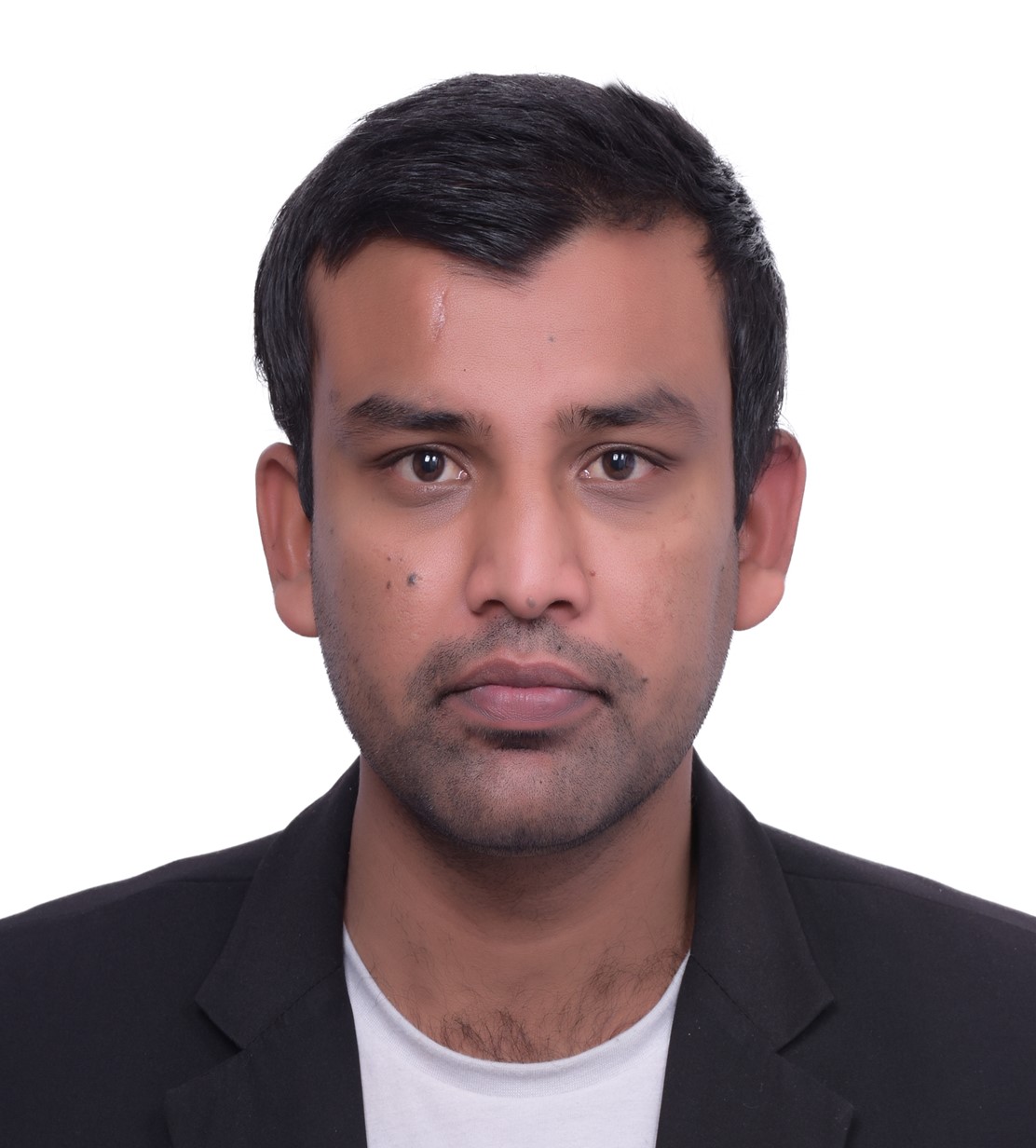}}]{Waleed Raza}received a B.E. degree in Electronic Engineering from the  Department of Electronic Engineering, Dawood University of Engineering and Technology Karachi, Pakistan in 2017. He received an M.S. degree in underwater acoustic communication engineering from the College of underwater acoustic engineering, Harbin Engineering University, Harbin China where he researched OFDM communication for underwater technology. He holds the editorial board member for Engineering, Technology, and Applied Science Research (ETASR) (2021-present), He is an active reviewer of a few journals including IEEE Sensor Journal, IEEE Access, and International Journal of Electronics and Communications (2018-2022), he has recently joined the IEEE as a student member. Currently, he is pursuing a Ph.D. degree at Embry Riddle Aeronautical University in Electrical and Computer Engineering. His research area of interest includes underwater acoustic OFDM communication, underwater acoustic target detection, artificial intelligence, machine learning for communication engineering, and autonomous unmanned systems such as UAVs and their characteristics. 
\end{IEEEbiography}
    
\begin{IEEEbiography}[{\includegraphics[width=1in,height=1.25in,,clip,keepaspectratio]{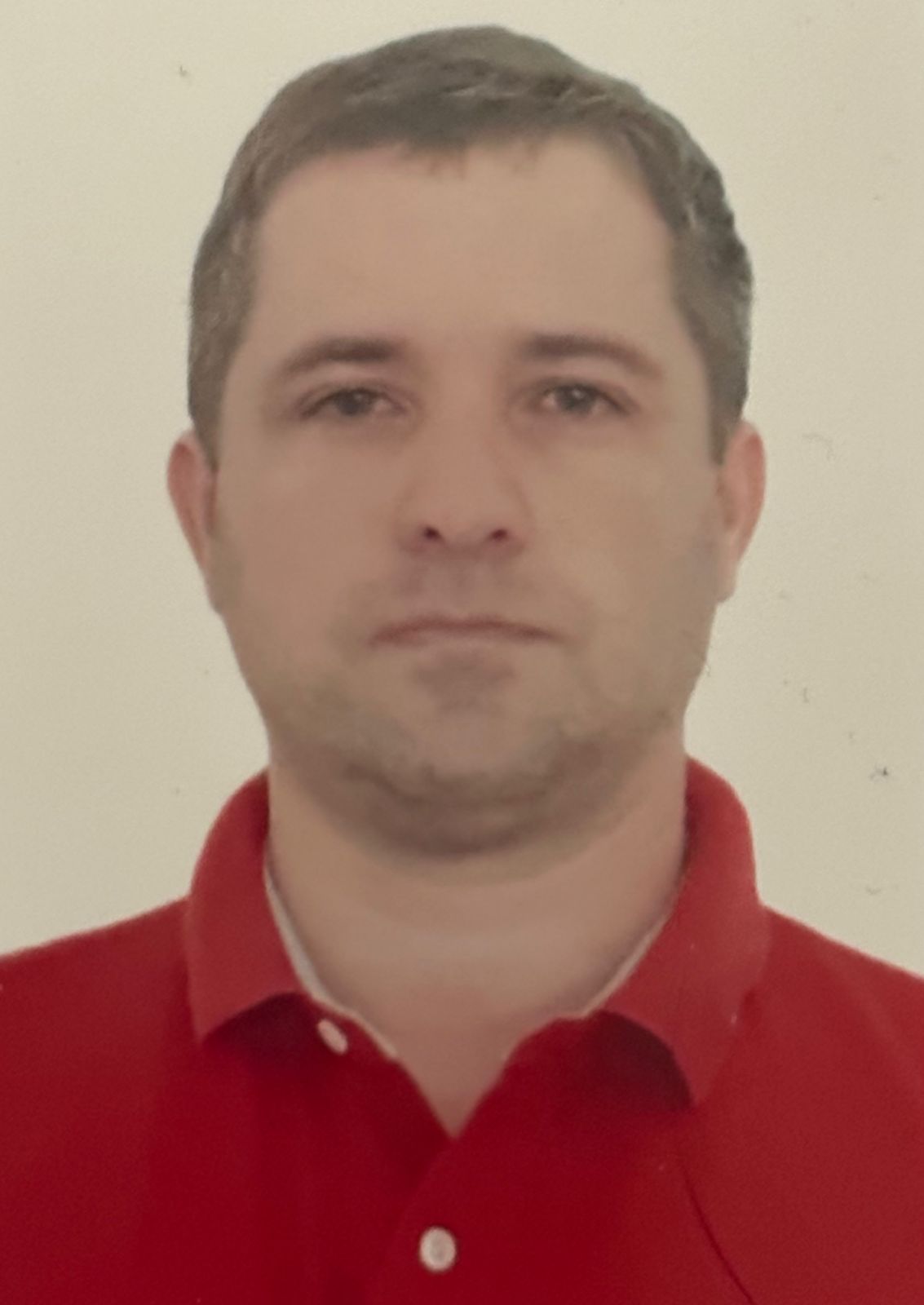}}]{Carlos M. J. M. Dourado Jr}{\space} received the Ph.D. degree in Informatics from the University of Fortaleza, Ceara, Brazil in February 2019 and MSc in Teleinformatics Engineering from the PPGETI/UFC (UFC, 2008). He completed a BSe in Electronics Engineering at the University of Fortaleza (Unifor, 2004). He is an associate professor and researcher at the Department of Telematics (DTEL)/Graduate Program in Computer Engineering (PPGCC) at the Federal Institutte of Ceara (IFCE), Brazil and PostDoctoral Research Associate in Embry-Riddle Aeronautical University. His main research areas include Internet of Things and Artificial Intelligence.
\end{IEEEbiography}

\begin{IEEEbiography}
[{\includegraphics[width=1in,height=1.25in,,clip,keepaspectratio]{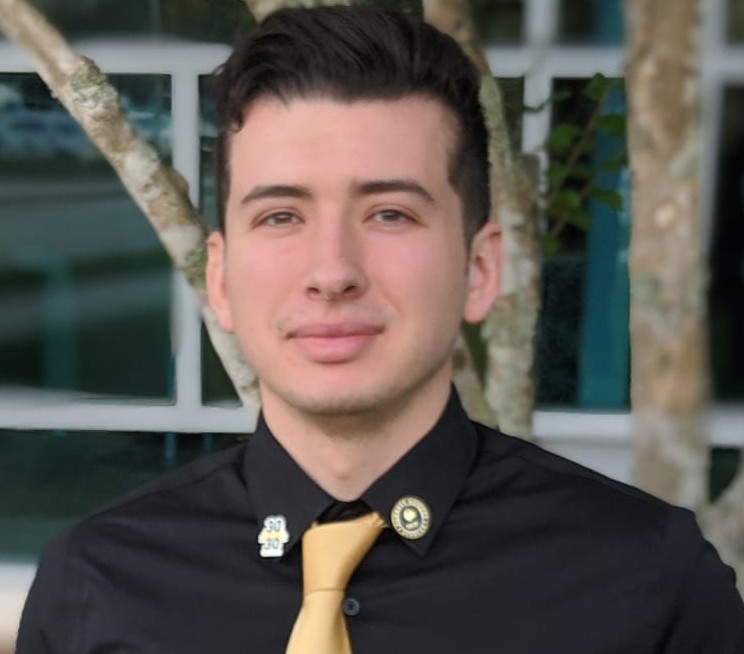}}]{Alvaro Velasquez} is a program manager in the Innovation Information Office (I2O) of the Defense Advanced Research Projects Agency (DARPA), where he currently leads the Assured Neuro-Symbolic Learning and Reasoning (ANSR) program. Before that, Alvaro oversaw the machine intelligence portfolio of investments for the Information Directorate of the Air Force Research Laboratory (AFRL). Alvaro received his PhD in Computer Science from the University of Central Florida and is a recipient of the National Science Foundation Graduate Research Fellowship Program (NSF GRFP) award, the University of Central Florida 30 Under 30 award, and best paper and patent awards from AFRL. He has co-authored 60 papers and two patents and serves as Associate Editor of the IEEE Transactions on Artificial Intelligence and his research has been funded by the Air Force Office of Scientific Research.
\end{IEEEbiography}

\begin{IEEEbiography}
[{\includegraphics[width=1in,height=1.25in,,clip,keepaspectratio]{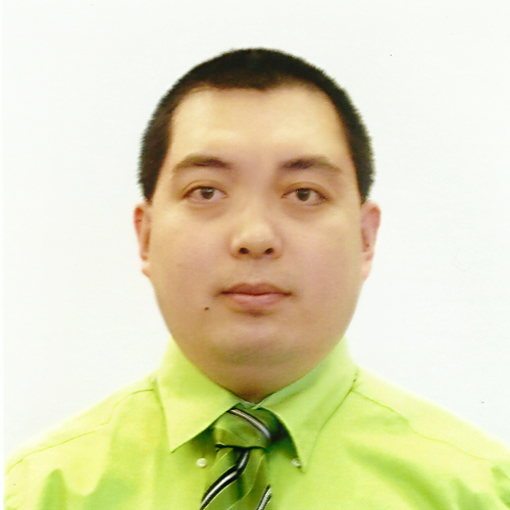}}]{Houbing Song} (M’12–SM’14-F’23) received the Ph.D. degree in electrical engineering from the University of Virginia, Charlottesville, VA, in August 2012.

He is currently a Tenured Associate Professor, the Director of NSF Center for Aviation Big Data Analytics (Planning), the Associate Director for Leadership of the DOT Transportation Cybersecurity Center for Advanced Research and Education (Tier 1 Center), and the Director of the Security and Optimization for Networked Globe Laboratory (SONG Lab, www.SONGLab.us), University of Maryland, Baltimore County (UMBC), Baltimore, MD. Prior to joining UMBC, he was a Tenured Associate Professor of Electrical Engineering and Computer Science at Embry-Riddle Aeronautical University, Daytona Beach, FL. He serves as an Associate Editor for IEEE Transactions on Artificial Intelligence (TAI) (2023-present), IEEE Internet of Things Journal (2020-present), IEEE Transactions on Intelligent Transportation Systems (2021-present), and IEEE Journal on Miniaturization for Air and Space Systems (J-MASS) (2020-present). He was an Associate Technical Editor for IEEE Communications Magazine (2017-2020). He is the editor of eight books, the author of more than 100 articles and the inventor of 2 patents. His research interests include cyber-physical systems/internet of things, cybersecurity and privacy, and AI/machine learning/big data analytics. His research has been sponsored by federal agencies (including National Science Foundation, US Department of Transportation, and Federal Aviation Administration, among others) and industry. His research has been featured by popular news media outlets, including IEEE GlobalSpec's Engineering360, Association for Uncrewed Vehicle Systems International (AUVSI), Security Magazine, CXOTech Magazine, Fox News, U.S. News \& World Report, The Washington Times, and New Atlas. 

Dr. Song is an IEEE Fellow and an ACM Distinguished Member. He has been an ACM Distinguished Speaker (2020-present), an IEEE Vehicular Technology Society (VTS) Distinguished Lecturer (2023-present) and an IEEE Systems Council Distinguished Lecturer (2023-present). Dr. Song has been a Highly Cited Researcher identified by Clarivate™ (2021, 2022). Dr. Song received Research.com Rising Star of Science Award in 2022, 2021 Harry Rowe Mimno Award bestowed by IEEE Aerospace and Electronic Systems Society, and 10+ Best Paper Awards from major international conferences, including IEEE CPSCom-2019, IEEE ICII 2019, IEEE/AIAA ICNS 2019, IEEE CBDCom 2020, WASA 2020, AIAA/ IEEE DASC 2021, IEEE GLOBECOM 2021 and IEEE INFOCOM 2022.
\end{IEEEbiography}

\end{document}